\documentclass[runningheads]{llncs}

\usepackage{eccv}

\usepackage{eccvabbrv}

\usepackage{graphicx}
\usepackage{booktabs}
\usepackage{multirow}

\usepackage[accsupp]{axessibility}  

\usepackage{comment}
\usepackage{sidecap}

\usepackage{hyperref}

\usepackage{orcidlink}

\begin{document}

\title{Commonly Interesting Images} 


\author{Fitim Abdullahu\orcidlink{0009-0008-9044-1868} \and
Helmut Grabner\orcidlink{0000-0003-1377-2686}}

\authorrunning{F. Abdullahu and H. Grabner}

\institute{School of Engineering, ZHAW, Zurich University of Applied Sciences \newline
\email{\{fitim.abdullahu, helmut.grabner\}@zhaw.ch}}

\maketitle

\begin{abstract}
Images tell stories, trigger emotions, and let us recall memories -- they make us think. Thus, they have the ability to attract and hold one's attention, which is the definition of being ``interesting''. Yet, the appeal of an image is highly subjective. Looking at the image of my son taking his first steps will always bring me back to this emotional moment, while it is just a blurry, quickly taken snapshot to most others. Preferences vary widely: some adore cats, others are dog enthusiasts, and a third group may not be fond of either. We argue that every image can be interesting to a particular observer under certain circumstances. This work particularly emphasizes subjective preferences. However, our analysis of $2.5k$ image collections from diverse users of the photo-sharing platform Flickr reveals that specific image characteristics make them \emph{commonly} more interesting. For instance, images, including professionally taken landscapes, appeal broadly due to their aesthetic qualities. In contrast, subjectively interesting images, such as those depicting personal or niche community events, resonate on a more individual level, often evoking personal memories and emotions.
  \keywords{Visual Interestingness \and Visual Attention}
\end{abstract}

\section{Introduction}
\label{sec:intro}
Over the past decade, our society has witnessed a remarkable shift towards visual communication, where visual imagery has taken center stage as a primary means of conveying information and messages \cite{Machin2014Handbooks4}. The highlighted prevalence of visual imagery invites us to explore a profound but often overlooked dimension - how we perceive what is considered \emph{interesting}. Visual interestingness, in essence, refers to an image's capacity to capture and retain an individual's attention \cite{Silvia2005WhatInterest.}. This trait holds tremendous importance as attention is the gateway to persuasion \cite{McGuire1968PersonalityTheory}. To influence attitudes, decisions, and behaviors, it is imperative that people first engage with a stimulus, including visual imagery \cite{Seo2013ThePersuasion}.
\begin{figure}[t]
    \centering
    \includegraphics[width=0.68\linewidth]{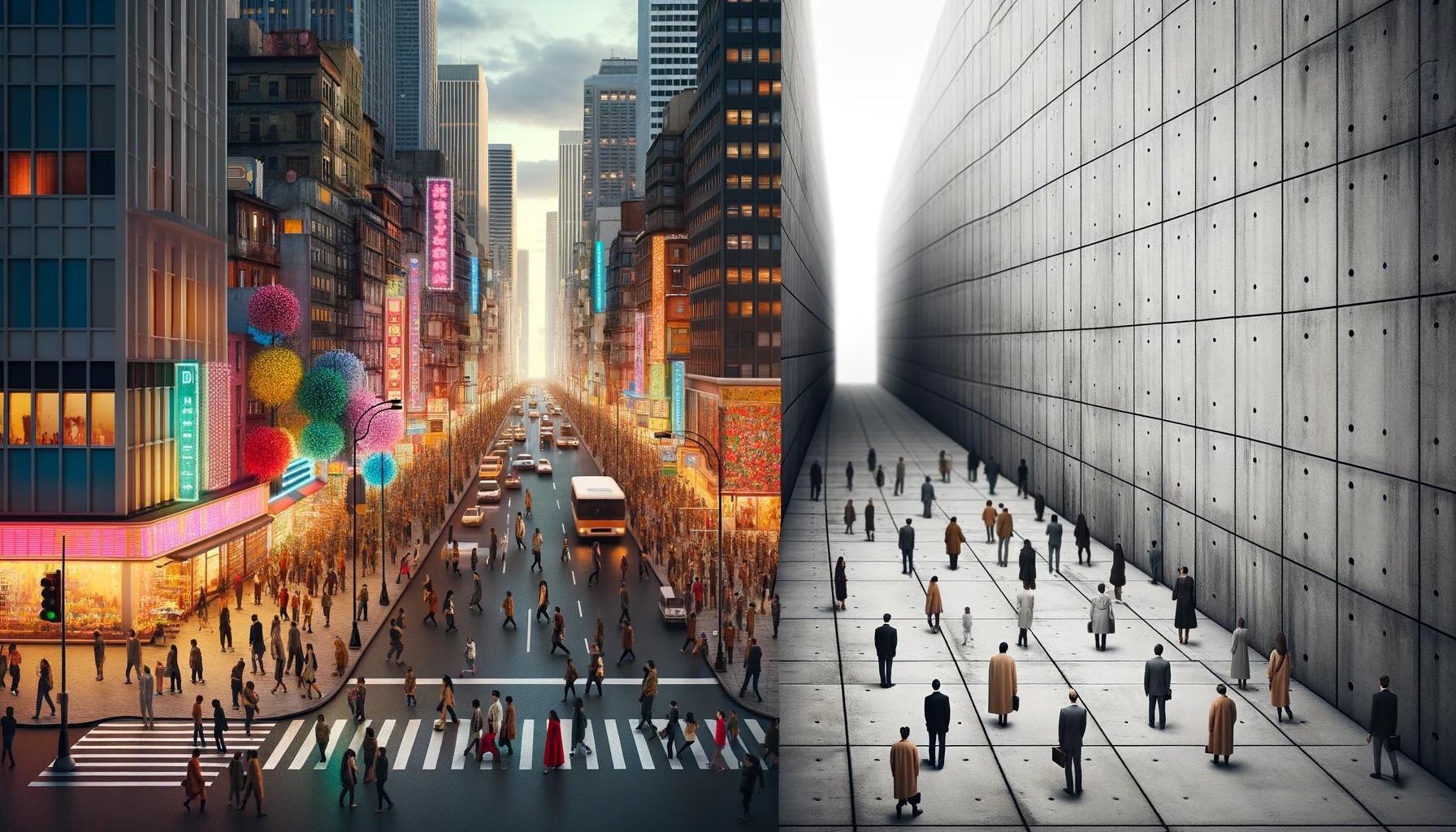}
    \caption{DALL·E 3: ``Give me an image which shows the contrast between interesting and uninteresting [...]'' ~\cite{dalle32023interesting}. In the generated image, many colors are used for the \emph{interesting} part, which is usually overfilled with objects and is generally quite complex. However, the \emph{uninteresting} part is usually depicted with few objects, monotonous and colorless, which gives a tendency towards simplicity.}
    \label{fig:DALL-E}
\end{figure}

Generative artificial intelligence models such as OpenAI's GPT-4V(ision)~\cite{openai2023gpt4, openai2023gpt4vision} or DALL·E 3~\cite{openai2023dalle3} create and analyze complex and engaging images, have further revolutionized visual communication and added a new dimension to the way we interact with and interpret images. Fig.~\ref{fig:DALL-E} depicts a generated image \cite{dalle32023interesting} when asking for an interesting image in contrast to a less interesting one. \emph{Interesting} seems to be usually used to describe a colorful image with many objects, leading to a tendency towards complexity. Uninteresting seems associated with monotony, colorlessness, and a trend towards simplicity. But aren't such images ``interesting'' in their own way? 

As has been repeatedly shown, \emph{interest} depends on the observer~\cite{Berlyne1949InterestConcept, Constantin2021VisualReview, schmidhuber2008driven}, \ie, it is \emph{subjective}. However, up to our knowledge, we are the first to make this subjectivity tangible. Our data-driven approach offers insights into commonly and subjectively interesting images, removing the strict distinction between interesting and uninteresting. We argue that every (!) image can be interesting to a particular observer under certain circumstances.

The remainder of the paper is organized as follows. Sec.~\ref{sec:related} briefly reviews related work. Sec.~\ref{sec:approach} introduces the \emph{FlickrUser}-dataset, which is used to define the \emph{common interest} (CI) of images. It turns out that certain semantic image categories appeal to many people (\ie, are of common interest). In contrast, other categories appeal to much smaller communities (\ie, are of subjective interest but still interesting). Sec.~\ref{sec:genInt} draws comparisons to different concepts of visual interestingness, aiming for a deeper understanding of the concept. Sec.~\ref{sec:compModel} presents results from a trained computational model based on our data-driven definition and discusses its limitations. Finally, Sec.~\ref{sec:conclusion} concludes the paper and outlines further work.

Our work will pave the way for a better understanding of the fuzzy concept ``interestingness'' by making the following main contributions:
\begin{itemize}
    \item We introduce the \emph{FlickrUser}-dataset containing $500k$ images from close to $2.5k$ users of the popular photosharing platform Flickr.
    \item Our analysis reveals factors contributing to common and subjective interest in images, including perceptual, denotative, and connotative features.
    \item Rather than a hard, interesting/ uninteresting definition, we propose a continuum ranging from common interestingness to very subjective interestingness, which is used to train a computational model.
\end{itemize}

\section{Related Work}
\label{sec:related}
A recent survey by Constantin et al.~\cite{Constantin2019ComputationalCovariates} provides a comprehensive overview of the methodologies, algorithms, and datasets employed in studying visual interestingness, underscoring the intricate and multifaceted nature of the subject. What makes an image interesting is of importance for various applications, such as event spotting~\cite{Grabner2013VisualSequences}, video summarization~\cite{Gygli2015VideoObjectives}, photo enhancement~\cite{bakhshi2019filtered}, to support people to organize and easily access their photo collection~\cite{photoColl, Dhar2011HighInterestingness} or marketing and advertisement~\cite{Schnurr2017}.

\textbf{Cognitive research.}
Work by Berlyne \cite{Berlyne1949InterestConcept} as early as 1949 ventured into the psychology of interestingness, shedding light on its subjective and dynamic nature. Berlyne's influential theory brought multifaceted aspects of interestingness to the forefront. He identified novelty, complexity, uncertainty, and conflict as crucial drivers of interestingness \cite{Berlyne1960ConflictCuriosity., Berlyne1970NoveltyValue}. These variables suggest that interest is generated in the human brain by comparing incoming information with the pre-existing knowledge of an observer. Individuals possess varying perspectives based on their experience and expertise, leading to distinct image processing in each person's brain -- being subjective. This subjective nature was also demonstrated in the recent work of Constantin et al.~\cite{Constantin2021VisualReview} in which human rates (only) moderately agree (Randolph's Kappa of $0.556$) on the task at hand.

While research on interestingness has made considerable strides, it has predominantly revolved around understanding which visual stimuli captivate human attention and the mechanisms underlying this captivation \cite{Tsotsos2005AAttention}. In the domain of cognitive psychology, it has been revealed that visual interestingness is shaped by two complementary forces: stimulus-based or bottom-up processing and memory-based top-down processing \cite{Shiffrin1977ControlledTheory.}. The intrinsic properties of an image influence bottom-up processing, while top-down processing is guided by factors of the reception situation, such as individual interests and goals. Bottom-up processing is more data-driven and less influenced by our expectations or previous experiences, allowing for a direct examination of what makes an image interesting. 

\emph{Bottom-up cues.} Bottom-up factors denote all features that are embedded within the image itself. These intrinsic image features include (a) perceptual, (b) denotative, and (c) connotative 
features~\cite{Barthes1977ImageText, Besson2003ADescription, Minu2014SemanticCreation}. Perceptual image features refer to basic syntactic properties, including color, contrast, quality, perspective, or composition. Denotative features describe the literal meaning of a depicted object, such as objects that are identified as representing a person, a dog, a chair, or a tree. Connotative features refer to the emotional or social association that a depicted object evokes. An example would be the depiction of a rose symbolizing love and affection.

\emph{Top-down cues.} As reviewed above, an image's interestingness does not solely depend on its intrinsic qualities. Take, for example, Kazimir Malevich's \emph{Black Square}~\cite{shatskikh2012black}. Everyone can easily draw a black square nowadays. Also, people before Malevich have drawn black squares, we are convinced. But only in 1915, in the right context and for a certain group of people, was it considered ``interesting'', finally becoming part of art history. Although the painting is visually simple, its philosophical context and historical relevance provide a deep canvas for top-down interpretation informed by the viewer's experience and knowledge. We can only engage with an image's interesting aspects if we grasp its motive and intent~\cite{Jia_2021_CVPR}. This interplay between bottom-up and top-down processes has been a central focus of investigation~\cite{Torralba2006ContextualSearch.}.

\textbf{Social- and visual interestingness.}
\emph{Visual} interestingness, as briefly reviewed above, encompasses elements like uniqueness, aesthetics, and subjective preferences related to depicted scenes, whereas \emph{social} interestingness is intertwined with the dynamics of social media platforms, including concepts like popularity, virality, and metrics such as views, likes, and shares \cite{deza2015virality}. It is extremely difficult to define visual interestingness based on the number of views or likes, mainly because the distributions of views and favorites are extremely skewed, \ie, almost all images have very few views or favorites. Whereas many views and likes point towards being interesting (due to the user's engagement), having an image with no views or likes does not imply an uninteresting image. See also our analysis in Sec.~\ref{sec:socialInt}.

The most related work might probably be that of Gygli et al.~\cite{Gygli2013TheImages} in which the authors define visual interestingness based on features/ cues they consider most important for capturing interest. Their predictor is then assessed with human-labeled data. Our approach defines visual interestingness directly based on user data, allowing us to learn the notion of interestingness rather than relying on predefined features.

\section{Data-driven Definition of Common Interest}
\label{sec:approach}
Based on image collection from many different Flickr users (Sec.~\ref{subsec:dataset}), we define \emph{common} interest (Sec.~\ref{subsec:ci}) and interpret results qualitatively (Sec.~\ref{subsec:interpretation}).

\subsection{The \emph{FlickrUser}-Dataset\protect\footnote{\url{https://github.com/fiabdu/Commonly-Interesting-Images}}} 
\label{subsec:dataset}

We chose Flickr as the source for our dataset. Flickr users can share and explore billions of images. These images are favored by diverse communities, including professionals and everyday users, representing varying common and subjective interest levels. Flickr images also serve for other datasets (such as Google Open Images~\cite{OpenImages2}), allowing us to augment and merge these sets to gain additional insights. Traffic demographics indicate that most users are from the US, followed by users from Europe (specifically the UK and Germany). The user base consists of approximately 40\% females and 60\% males, with most users falling within the 25 to 34 age range \cite{similarweb}.

For a given user, we download publicly shared images it likes. We randomly chose $2,337$ unique users with at least $10$ images liked each. For computational reasons, we randomly draw a maximum of $1,000$ images per user, finally leading to $504,241$ images. We posit that all these images inherently possess some interestingness based on specific, maybe subjective, characteristics; otherwise, they would not have been captured, uploaded, and later liked by users~(\cf~\cite{McGuire1968PersonalityTheory}).

\subsection{Common Interestingness (CI)}
\label{subsec:ci}
Our approach explores the nuances of visual interestingness by scoring images as either commonly or subjectively interesting. In contrast to defining interestingness solely based on metadata attributes, we based our definition and analysis on image collections by different users. The main idea is to identify semantically similar images that different users like. If many unique users like a certain type of image, this type of image is considered to be more of a common interest. Consequently, if a certain kind of image is liked only by a few users, this type is less common and more subjectively interesting.

\begin{figure}[ht!]
    \centering
    \begin{subfigure}{\linewidth}
        \includegraphics[width=1\linewidth]{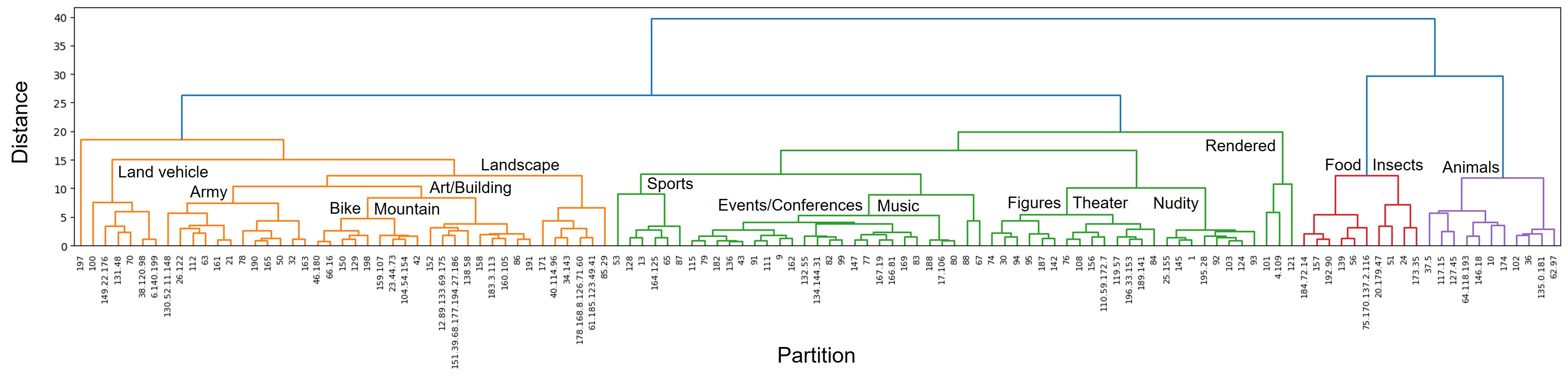}
        \caption{Semantic clustering of the CLIP~\cite{radford2021learning} space into non-overlapping partitions. As can be seen, specific topics are consecutively merged into more generic categories. Each partition relates to a semantic concept, building the basis for our analysis.}
        \vspace{1em}
        \label{fig:hierarchicalMerged}
    \end{subfigure}
    
    \begin{subfigure}{\linewidth}
        \includegraphics[width=1\linewidth]{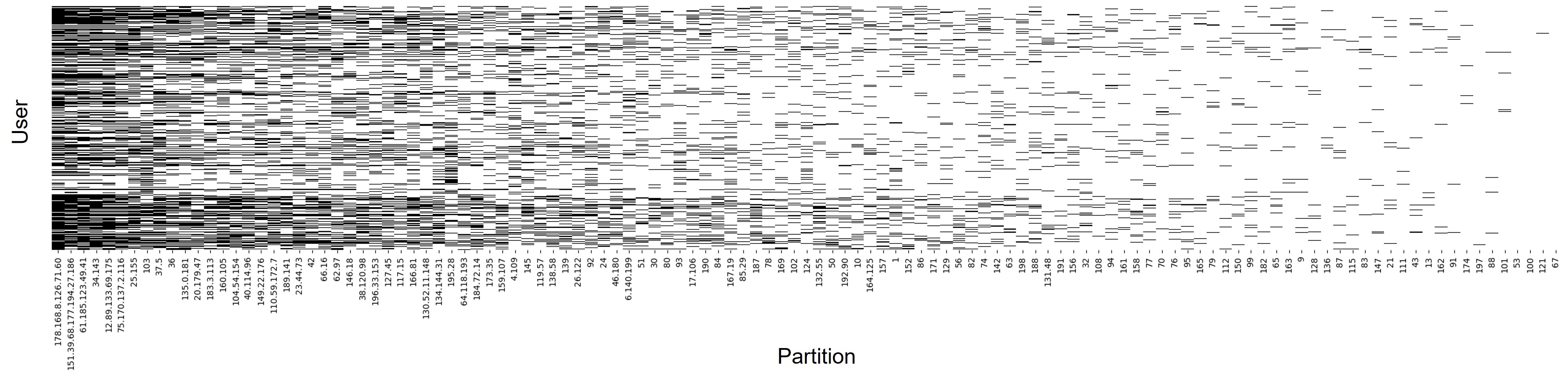}
        \caption{Liked images per user in every partition (black: user has at least one image in that partition; white: otherwise). Users can have images in as many partitions as possible, but only one image counts within one partition. Sorted by the count in each partition to align with (c).}
        \vspace{1em}
        \label{fig:heatmapMerged}
    \end{subfigure}
    
    \begin{subfigure}{1\linewidth}
        \includegraphics[width=1\linewidth]{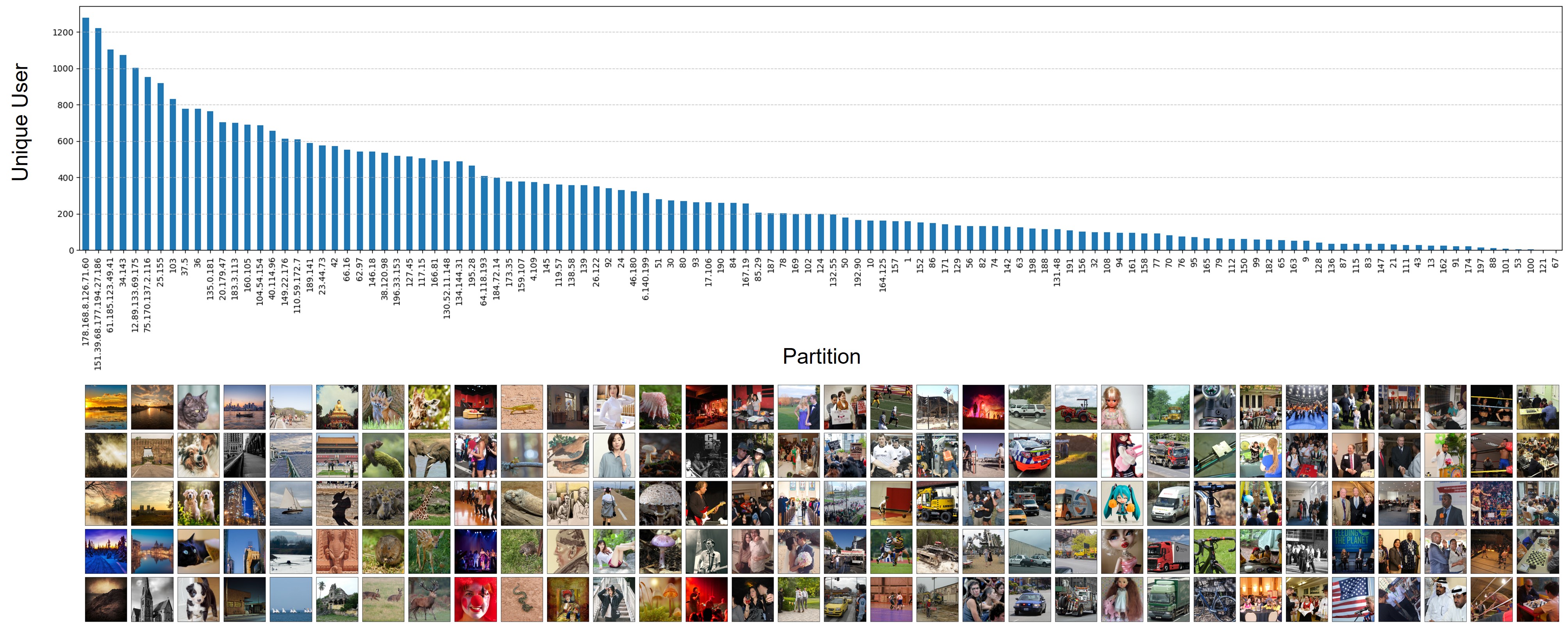}
        \vspace{1em}
        \caption{The sorted arrangement of partitions according to the $CI$ scores (proportional to the unique users displayed here) offers insights into common and subjective image interestingness. Commonly interesting images include aesthetic landscapes, black and white images that evoke emotions, images of buildings from a particular perspective, and images with little content, such as raindrops on a window. More subjective image sets show various topics, such as rendered images of the video game Second Life, different sports, animals, toys, and people that show a lot of skin and are sexually provocative. 
    }
        
        \label{fig:ourInterestingness}
    \end{subfigure}
    \caption{The image space is partitioned into semantic similar partitions using k-means and hierarchical clustering (a). Images from users are assigned to their respective partitions (b). The more users have liked an image in a particular partition, the more \emph{commonly} interesting it is considered (c). {\color{blue}{All partitions, $CI$ scores, and typical images are included in the supplementary material for a closer investigation.}}
    }
    \label{fig:vis}
\end{figure}
Our proposed approach, described in the following, is quite straightforward yet effective.

{\bf Image space partitioning.} To get a semantic description of an image, we use features from the CLIP (Contrastive Language–Image Pre-training) model ViT-L/14@336px ~\cite{radford2021learning}, which have shown impressive results for various computer vision tasks lately. We partition the feature space with k-means into $N$ partitions. To achieve a more stable partitioning but still maintain the local structure of the data, we first applied UMAP~\cite{mcinnes2018umap} to reduce the dimension from 768 (CLIP embedding) down to 7 (empirically estimated) \cite{umaplearn}.

{\bf Unique users per partition.} Images that a user has liked are assigned to the individual partitions. The more unique users have images in a particular partition, the more agreement they have about liking them -- the more commonly interesting the type of images in that partition are. More formally, let $p_i$ be a partition reflection of a certain semantic type of images, the set of unique users
\begin{equation}
  UU(p_i) = \{ user\ |\ \exists_k: favImg^{(user)}_k \in p_i\} 
  \label{eq:uniquUsers}
\end{equation}
consist of all users $user$, which have at least liked one\footnote{Increasing the necessary number of likes per user will decrease the number of unique users per partition, as will the $CI$ score. This does not significantly affect partitions with higher $CI$ scores, but partitions with lower $CI$ scores disappear, indicating poor capture of subjectivity.} image $favImg^{(user)}_k$ falling into the partition $p_i$. The common interest
\begin{equation}
  CI(p_i) = \frac{1}{M} \cdot | UU (p_i) |
  \label{eq:commonInterest}
\end{equation}
is defined as the number of unique users normalized by the total number of users $M$.

\begin{figure}[tb!]
  \centering
   \includegraphics[width=1\linewidth]{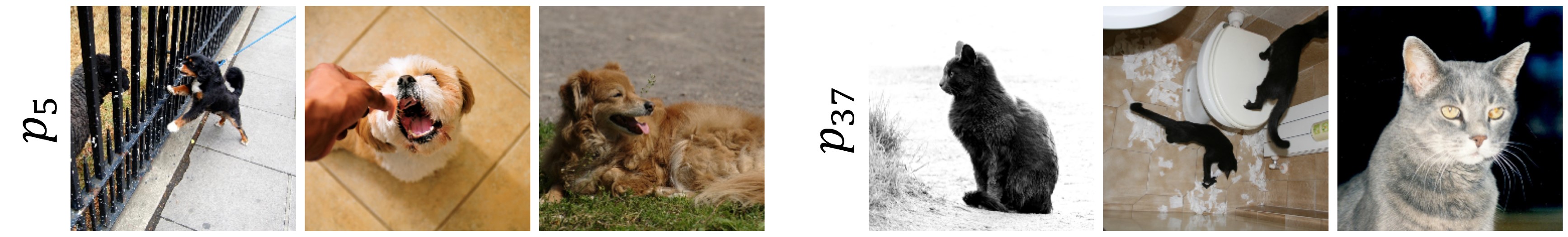}
   \caption{Example of a merged cluster: Around 60 percent of users who prefer cat images in partition $p_{37}$ also prefer dog images in partition $p_{5}$. As the semantic similarity $d(p_{37},p_{5}) = 1.7$ is close enough these two partitions get merged into $p_{37.5}$.}
   \label{fig:mergedClusterExample}
\end{figure}

{\bf Refining image space partitioning.}
The definition of $CI$  allows us to merge similar -- concerning image similarity and similar $CI$ score -- partitions. In fact, we apply bottom-up hierarchical clustering to recursively merge the initial $N$ partitions to more general topics. Two partitions $p_i$ and $p_j$ are merged into partition $p_{ij} = p_i \cup p_j$ if the image similarity, expressed by the Ward distance $d(p_i, p_j) < \theta_{image}$ used for the hierarchical clustering and the common interest of the two partitions is similar as well, \ie, the intersection over union (IoU) satisfies $\frac{UU(p_i) \cap UU(p_j)}{UU(p_i) \cup UU(p_j)} > \theta_{CI}$. Both parameters are estimated experimentally and set to $\theta_{image} = 3$, $\theta_{CI} = 0.25$. Merging is repeated until convergence.\footnote{Please note that the initial selection of $N$ clusters for the k-means partitioning might have seemed arbitrary. However, one has to ensure that it is ``fine'' enough to capture all topics and, simultaneously, large enough to allow for a robust estimation of $CI(\cdot)$. We got good results with $N \in [150, 300]$, where we chose $N=200$ for the rest of the study.} In this way, we ensure that similar clusters with a certain proportion of identical users are merged; see Fig.~\ref{fig:mergedClusterExample}. We ended up with $119$ partitions.

\begin{figure}[tb]
    \centering
    \begin{subfigure}{0.45\linewidth}
        \includegraphics[width=1\linewidth]{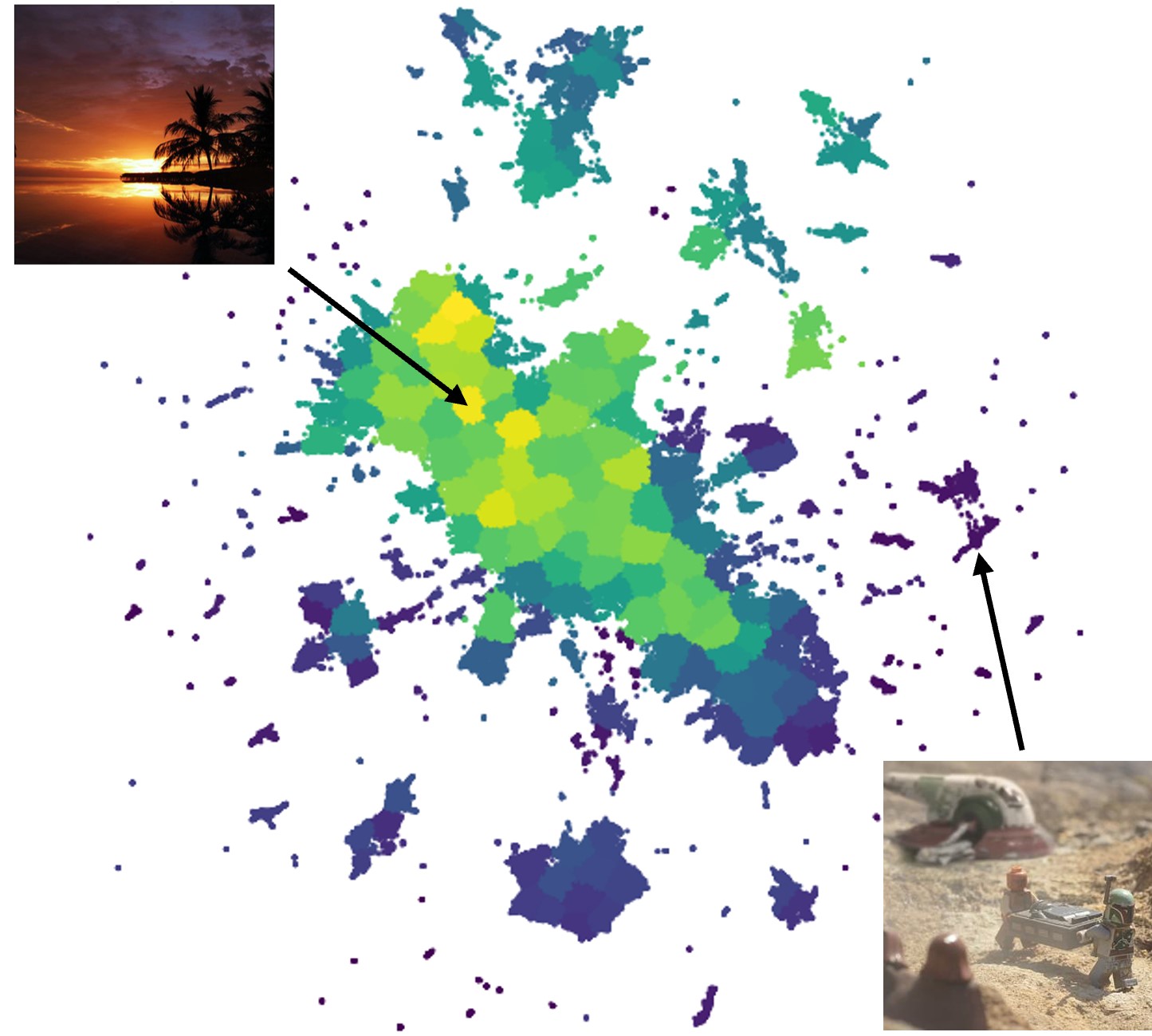}
        \caption{2d-projection of the partitions, colored according to the proposed \emph{common} interest score $CI$. Yellow: commonly interesting; purple: subjectively interesting.}
        \label{fig:partitionPlot}
    \end{subfigure}
    \hfill
    \begin{subfigure}{0.5\linewidth}
        \includegraphics[width=1\linewidth]{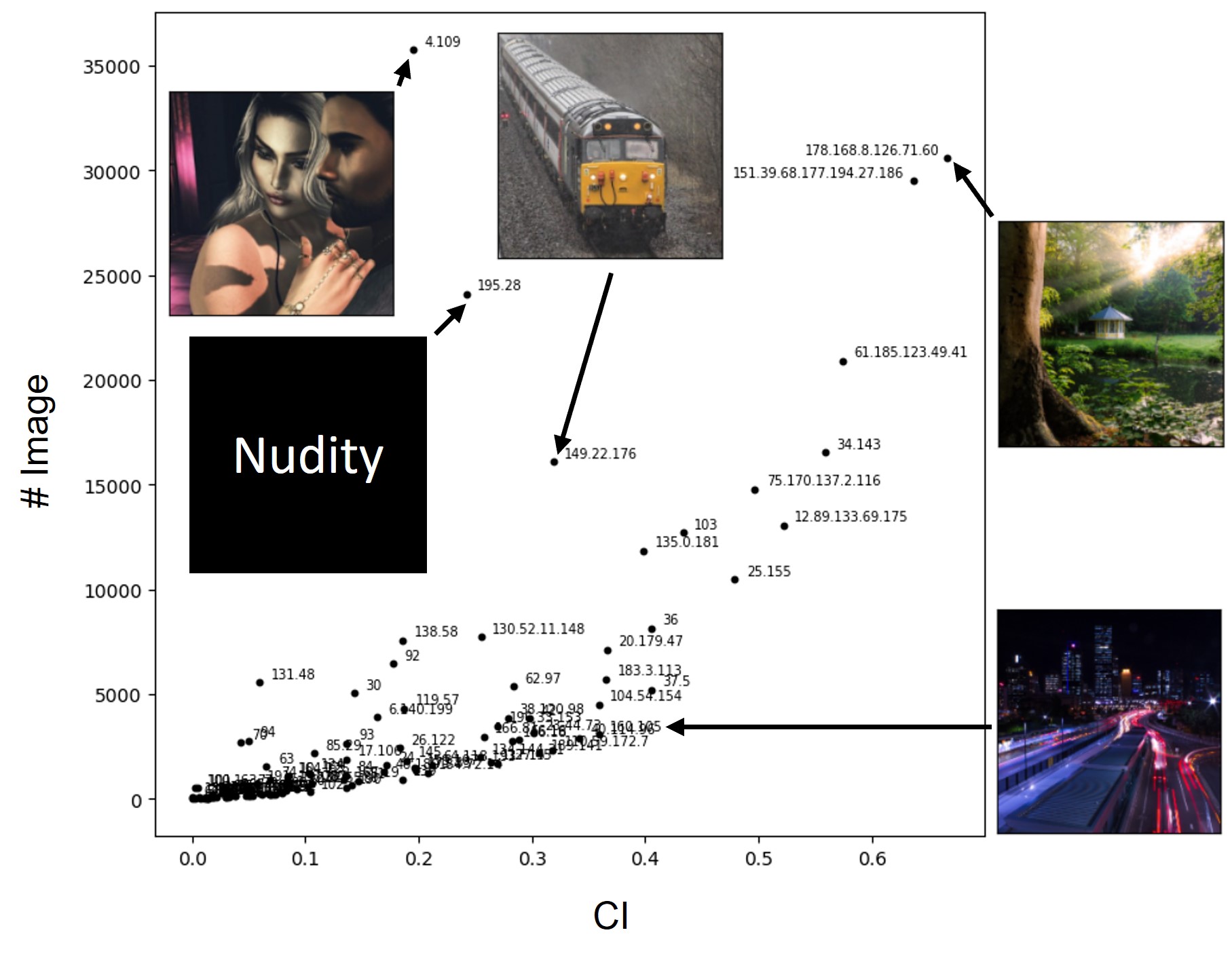}
        \caption{The number of images in each partitioning must not necessarily correlate to their common interestingness. Several exceptions exist, \eg, Second Life, nudity, passionate train lovers, \etc.}
        \label{fig:imgCountCompare}
    \end{subfigure}
    \caption{$CI$ scores overlaid on the image partitions space (a) and correlation with the number of images in each partition (b).
    }
\end{figure}

{\bf Visualization.}
Fig.~\ref{fig:vis} depicts the individual steps and results. This final partitioning after refining is shown as a dendrogram in Fig.~\ref{fig:hierarchicalMerged} and projected in 2d in Fig.~\ref{fig:partitionPlot}. Please note that semantic concepts have been nicely grouped. Fig.~\ref{fig:heatmapMerged} depicts details of user likes per partitioning, sorted by unique users. Similarly, the number of unique users, proportional to the $CI$ score, and typical images are shown in Fig.~\ref{fig:ourInterestingness}.

{\bf Number of images in each partition.} One might think that higher $CI$ scores imply many images in that partition. As depicted in Fig.~\ref{fig:imgCountCompare} this is only to some extent the case. Images of a very subjective nature might be from a smaller community and thus result in fewer uploads. Images of very high common interests might appear more frequently because photographers are motivated to produce more. However, they are overtaken by images from the video game Second Life, followed by many (questionable) images showing people with a lot of skin (\cf discussion at the end of the paper in Sec.~\ref{sec:conclusion}). Quantitatively, the median $CI$ score in the dataset is 0.32, approximately half of the maximum $CI$ score, indicating that the amount of data per partition is independent of the $CI$ score.

\subsection{Interpretation and Manual Analysis}
\label{subsec:interpretation}
Partitions with high $CI$ scores point towards commonly interesting images. For instance, over $65\%$ (\ie, $CI(p^\star)=0.65$) liked images in the first partition $p^\star$, making semantic similar images of this cluster commonly interesting -- including images showing aesthetic landscapes and forests, mostly with touching colors and beautiful sunsets. Conversely, those favored by a niche audience, \eg, Star Wars figurines $(p_{30})$, are considered subjectively interesting; $CI(p_{30})=0.14$. 

The lower the consensus among users -- the lower the $CI$ score -- the more subjectively interesting the images become. Examples are images of animals, professional paintings, and professional portraits of people. Subjectively interesting images include various sports such as cycling or wrestling, groups of people in conferences, and people playing chess. Some observations:\footnote{Please take it with a grain of salt and build your own opinion. All partitions, along with typical images, are included in the supplementary material.} Trains are more interesting than buses. Main courses are (slightly) more interesting than desserts, and fruits predominate for both dishes. Vintage cars are more interesting than modern cars. Bikes with people on them are more interesting than just bikes. Similarly, flowers with insects are more interesting than just flowers. LEGO is more interesting than dolls. There seems to be no difference between dogs and cats; still, they are more interesting than reptiles.

\section{What Makes an Image Commonly Interesting?}
\label{sec:genInt}
The findings from the previous section will be connected to topics such as social interestingness (Sec.~\ref{sec:socialInt}) and intrinsic image attributes (Sec.~\ref{sec:perceptualFeatures} to~\ref{sec:connFeatures}).

To ensure a uniform analysis, we divided the \emph{FlickrUser} dataset into three groups of equal size, sorted by cumulative $CI$ score from highest to lowest. It's important to note that this grouping is intended solely to discern trends in what factors contribute to an image's level of interest. We determined that three groups suffice for this objective. The groups representing images that are more commonly interesting (\emph{Comm.}; first 14 partitions), very subjectively interesting (\emph{Subj.}; last 83 partitions), and represent an interplay of both (\emph{Inter.}; 22 partitions in the middle).\footnote{Demographics: Approximately 64\% of the users have specified no gender. Male, Female, and Other are consistent across all groups: Male (26.68\% ± 0.40\%), Female (8.52\% ± 0.60\%), and Other (0.40\% ± 0.05\%). Users' place of residence is also consistent, with the top three: Pacific Time (34.80\% ± 0.23\%), GMT (12.66\% ± 0.11\%), and Eastern Time (11.83\% ± 0.08\%). The age of the users cannot be obtained. Our drawn conclusions are only marginally affected.}  

\subsection{Social Interestingness}
\label{sec:socialInt}

\begin{figure}[h]
    \centering
    \begin{subfigure}{\linewidth}
        \includegraphics[width=1\linewidth]{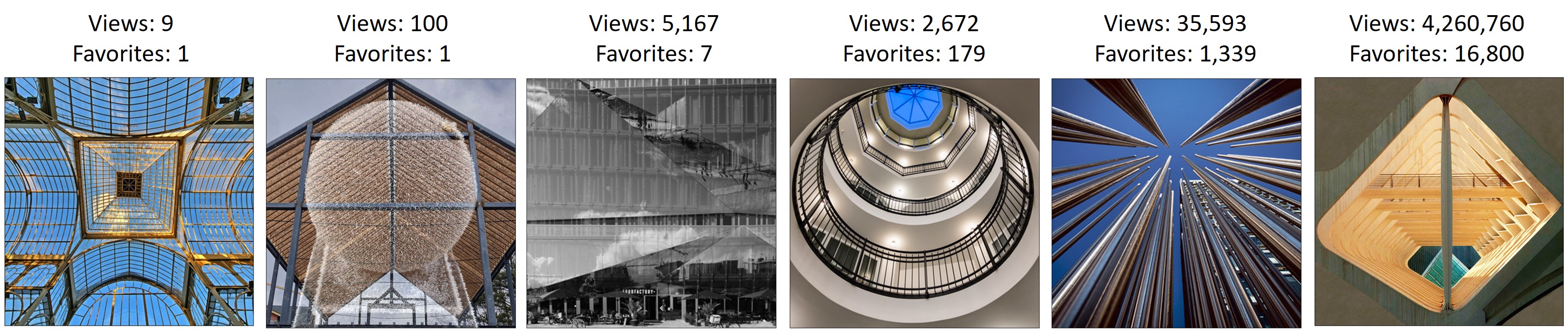}
        \caption{All these images are from the same partition and, as such, have the same $CI$ score of 0.64. Thus, they are equally interesting according to our definition. However, social interestingness differs significantly as the total number of views and likes (as well as their ratio) varies a lot.}
        \vspace{0.5em}
        \label{fig:viewsFavoritesCompare}
    \end{subfigure}
    \vfill
    \begin{subfigure}{\linewidth}
        \includegraphics[width=1\linewidth]{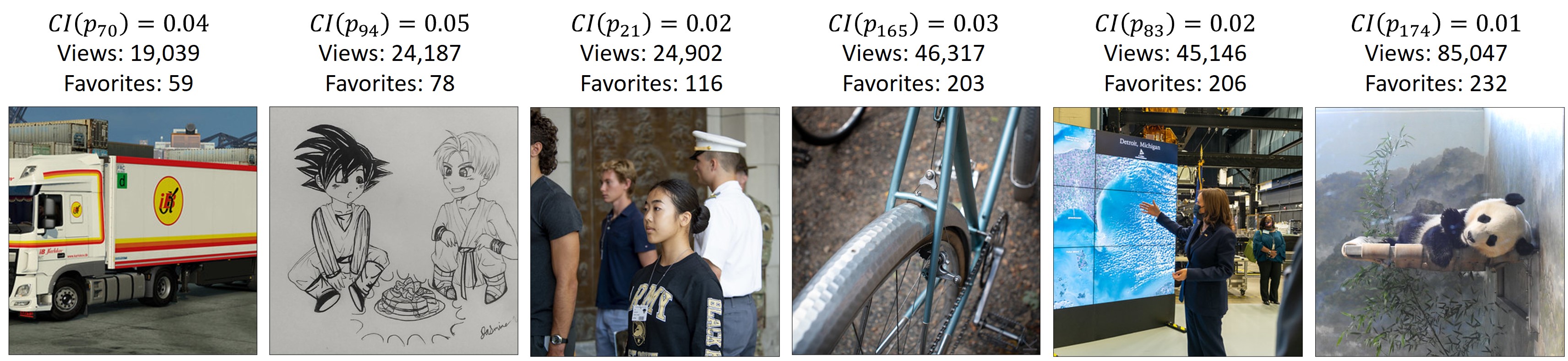}
        \caption{Many images from Flickr's  (social) ``interestingness'' category align with our definition of common interestingness, yet these examples demonstrate notable differences.}
        \label{fig:viewsFavoritesCompareTwo}
    \end{subfigure}
    \caption{Meta-data such as the \emph{absolute} number of views and likes as metrics are unreliable due to recommendation system influences. An image with a few likes or views can also be interesting (a), while viral images might be irrelevant to many users (b).
    }
    \label{fig:viewsFavorites}
\end{figure}

Using the absolute number of views and likes as a proxy of an image's interest is not a dependable approach due to the potential for recommendation systems to skew these measures heavily. To overcome these issues, it is worth remembering that our definition of visual interest is independent of the absolute number of likes. Examples for $CI(p_{151.39.68.177.194.27.186})= 0.64$ are given in Fig. \ref{fig:viewsFavoritesCompare}. While all images are deemed commonly interesting according to our definition, there is a significant variance in views and favorites.

{\bf Flickr's Interestingness.} In line with~\cite{Dhar2011HighInterestingness}, we employed images from Flickr's ``Interestingness'' category for our analysis~\cite{flickrPat, flickraboutint}. We aimed to compare social interestingness with our definition of interestingness. Therefore, we collected $149,457$ random images and assigned them to their corresponding partition. The majority of $77.05\%$ falls within the common interesting category, followed by $16.60\%$ for the intermediate and $6.35\%$ for the subjective category. Example images considered subjective by our approach are shown in Fig.~\ref{fig:viewsFavoritesCompareTwo}.

\subsection{Perceptual Features}
\label{sec:perceptualFeatures}

We used the Vision-Language Aesthetics (VILA) model~\cite{ke2023vila} to obtain perceptual feature scores. The model is pre-trained on aesthetic image comments from photography-sharing platforms, including Flickr, providing a reliable aesthetic quality score. Additionally, perceptual features and photographic styles such as complementary colors, vanishing points, and HDR are estimated using zero-shot learning~\cite{ke2023vila, murray2012ava}.
\begin{SCtable}
    \caption{Perceptual Features sorted by the highest differences $\Delta$ between the \emph{Comm.} and \emph{Subj.} groups. Although some features are present/ absent in both groups, especially those that correspond to image quality, they contribute to distinguishing between more or less commonly interesting images.}
    \label{tab:percTable}
    \centering
    \tiny
    \begin{tabular}{@{}p{2.5cm}|r|r|r|r@{}} 
        \toprule
        Scores from \cite{ke2023vila} & Comm. [\%] & Inter. [\%] & Subj. [\%] & $\Delta$\\
        \midrule
        \midrule
        HDR &  20.68 & { 15.81} & 15.22 & 5.46\\
        Shallow DoF & 8.39 & 4.51 & 3.10 & 5.29 \\
        Vanishing Point & 8.88 & 9.59 & 4.39 & 4.49\\
        Long Exposure & 6.68 & 1.52 & 2.42 & 4.26\\
        Silhouette & 3.51 & 1.33 & 0.87 & 2.64\\
        Macro & 3.76 & 1.65 & 2.06 & 1.70\\
        Soft Focus & 10.03 & 7.92 & 9.41 & 0.62\\
        Image Grain & 1.48 & 0.97 & 1.73 & -0.25\\
        Rule of Thirds & 14.82 & 23.92 & 15.43 & -0.61\\
        Light on White & 1.37 & 1.69 & 2.29 & -0.92\\  
        Motion Blur & 1.91 & 3.98 & 3.79 & -1.88\\
        Compl. Colors & 5.21 & 9.33 & 7.65 & -2.44 \\
        Duo Tones & 6.75 & 9.24 & 13.42 & -6.67\\          
        Negative Image & 6.50 & 8.54 & 18.22 & -11.72\\
        \midrule
        Photographer & 5.79 & 4.16 & 2.46 & 3.33\\
        \midrule
        \multicolumn{2}{l}{Aesthetic Score (higher: better)}\\
        \ \ \ $q_{25}$ & 47.03 & 39.30 & 38.14 \\
        \ \ \ $q_{50}$ (median) & 55.49 & 47.78 & 46.72 & 8.77 \\
        \ \ \ $q_{75}$ & 63.66 & 56.14 & 55.34 \\
        \bottomrule
    \end{tabular}
\end{SCtable}

As seen from Tab.~\ref{tab:percTable}, the common interest group contains many HDR images. This proportion decreases in groups that are more likely to be assigned to subjective interest. Images taken according to the Rule of Thirds are strongly represented in all three groups, which indicates that this characteristic says little about the common or subjective interest. As subjective interest increases, the proportion of negative images and motion blur increases; overall, the image aesthetic score decreases. Generally speaking, according to self-reported occupation labels, skilled or professional photographers seem to take commonly interesting images. This aligns with previous research, which found that aesthetic scores are higher for professional photographers \cite{strukova2024identifying}.

\subsection{Denotative Features}
\label{sec:denotativeFeatures}
We used images from the Google Open Images V7 database~\cite{OpenImages2}, also built on Flickr images, to retain the denotative features. We assigned a sample of $1,163,050$ images to our partitions. These images contain $3,275,047$ point labels (an image usually contains several point labels), providing information about the objects (person, cat, \etc) and stuff (sky, water, \etc) in an image \cite{benenson2022colouring}.

\begin{SCtable}
  \caption{Top 15 denotative features (ground-truth annotation from Google Open Images~\cite{OpenImages2} point labels). Common interest is mainly characterized by features corresponding to natural or city scenes. Although the label person is represented in all three groups, their proportion increases with increasing subjectivity (from 2.15\% to 5.57\%).}
  \label{tab:denTable}
  \centering
  \tiny
    \begin{tabular}{@{}p{1.5cm}p{0.5cm}|p{1.5cm}p{0.5cm}|p{1.5cm}p{0.5cm}@{}} 
      \toprule
      \multicolumn{2}{c|}{Comm. [\%]}  & \multicolumn{2}{c|}{Inter. [\%]}  & \multicolumn{2}{c}{Subj. [\%]} \\
      \midrule
      \midrule
      Flower & 5.69 & Clothing & 5.20 & Clothing & 7.53 \\
      Tree & 4.61 & Person & 3.88 & Person & 5.57 \\
      Dog & 2.95 & Car & 3.09 & Man & 4.99 \\
      Cloud & 2.92 & Man & 3.00 & Woman & 4.43 \\
      Car & 2.91 & Wheel & 2.35 & Girl & 3.19 \\
      Plant & 2.89 & Woman & 2.35 & Footwear & 2.62 \\
      Sky & 2.81 & Tree & 2.34 & Wheel & 2.48 \\
      Building & 2.69 & Table & 2.17 & Table & 2.21 \\
      Bird & 2.52 & Girl & 2.08 & Suit & 2.00 \\
      Person & 2.15 & Footwear & 1.77 & Car & 2.00 \\
      Skyscraper & 2.05 & Sky & 1.57 & Human body & 1.93 \\
      Clothing & 1.84 & Chair & 1.53 & Boy & 1.63 \\
      Grass family & 1.83 & Boat & 1.53 & Chair & 1.48 \\
      Cat & 1.73 & Wall & 1.41 & Dress & 1.42 \\
      Sculpture & 1.71 & Boy & 1.37 & Tree & 1.31 \\
      \bottomrule
    \end{tabular}
\end{SCtable}

Tab.~\ref{tab:denTable} shows the ranking of the sum of individual objects in the images. The common interest group is mainly characterized by denotative features usually associated with landscapes or cityscapes, such as flowers, sky, or buildings. Images of animals such as dogs, cats, or birds are also related to common interest. In contrast, the subjectively interesting group is characterized by denotative features such as clothing, person, or human body, typically associated with images of people. Cars are present in all three groups, but their most significant representation is in the interplay group.

\subsection{Connotative Features}
\label{sec:connFeatures}
Estimating emotions from images is a challenging problem. We decided to use the CLIP~\cite{radford2021learning} vision-language model to compare text prompts of emotions with images. In emotional studies~\cite{zhao2016predicting, iaps2010}, eight basic emotions aligned with keywords have been established. Four are positive (excitement, awe, amusement, and contentment), and four are negative (sadness, disgust, anger, and fear).

As seen in Tab.~\ref{tab:connTable}, images in the common interestingness group evoke excitement and awe. Funny images and images that make one happy are more likely to be found in the subjective group. About $89\%$ of the images in the common interestingness group evoke positive emotions. This number decreases as the group becomes more subjective. Consequently, the negative categories are also more likely to be found in the subjective category, but the proportion is minor in all three groups. 

\begin{SCtable}
  \caption{Connotative features sorted by the highest differences $\Delta$ between the \emph{Comm.} and \emph{Subj.} groups. Common interestingness evokes excitement associated with the words ``thrilling'' or ``astonishing''. Overall, the more subjective the images are, the higher the proportion of negative emotions.}
  \label{tab:connTable}
  \centering
  \tiny
    \begin{tabular}{@{}p{2.5cm}|r|r|r|r@{}} 
      \toprule
      CLIP scores~\cite{radford2021learning} & Comm. [\%] & Inter. [\%] & Subj. [\%] & $\Delta$\\
      \midrule
      \midrule
      Excitement \cite{clip2023excitement} & 60.79 & 35.58 & 32.14 & 28.65\\
      Awe \cite{clip2023awe} & 18.08 & 15.80 & 9.40 & 8.68 \\   
      Contentment \cite{clip2023contentment} & 4.79 & 10.79 & 9.43 & -4.64\\
      Amusement \cite{clip2023amusement} & 5.59 & 16.01 & 17.55 & -11.96 \\   
      \midrule
      Sum positive & 89.25 & 78.18 & 68.52 & 20.73\\
      \midrule
      \midrule   
      Fear \cite{clip2023fear} & 0.65 & 2.84 & 2.10 & -1.45\\
      Disgust \cite{clip2023disgust} & 0.91 & 3.81 & 6.60 & -5.69 \\
      Sadness \cite{clip2023sadness} & 8.25 & 12.33 & 14.63 & -6.38\\
      Anger \cite{clip2023anger} & 0.95 & 2.85 & 8.16 & -7.21 \\
      \midrule
      Sum negative & 10.76 & 21.83 & 31.49 & -20.73\\
      \bottomrule
    \end{tabular}
\end{SCtable}

\section{Computational Model of Common Interestingness}
\label{sec:compModel}
To evaluate an image $\mathbf{x}$, it will be first assigned to its partition $p_i$, and the corresponding $CI(p_i)$ score will be returned. The result will be very coarse as all images assigned to partition $p_i$ will have the same score. We trained a simple linear regression on the original 768-dimensional CLIP embeddings to obtain a more fine-grained measurement. As a target, the $[0, 1]$ normalized $CI$ score of the images in the respective partitions is used as they reflect our data-driven definition of common interestingness. The trained model $CI_R(CLIP(\mathbf{x}))$ can be applied directly to an image.\footnote{On an independent test set, a $R^2$ value of 0.66 was obtained.}
\begin{figure}[t]
  \centering
    \begin{subfigure}{\linewidth}
        \includegraphics[width=1\linewidth]{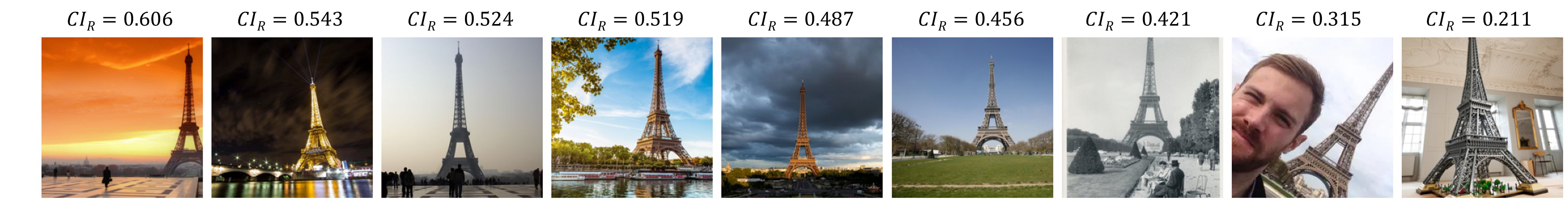}
        \caption{Eiffel Tower}
        \label{fig:eiffelTower}
    \end{subfigure}
    \vfill  
    \begin{subfigure}{\linewidth}
        \includegraphics[width=1\linewidth]{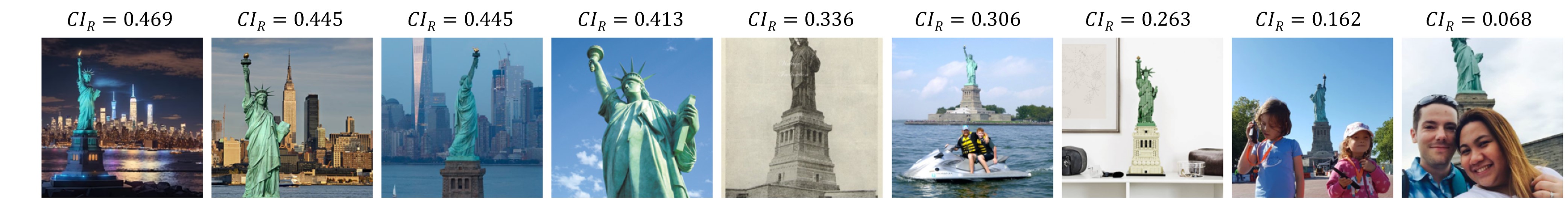}
        \caption{Statue of Liberty}
        \label{fig:statueOfLiberty}
    \end{subfigure}
    \caption{Rankings of images according to their common interestingness (high to low from left to right).}
\end{figure}

\begin{figure}[p]
    \centering
    \begin{subfigure}{\linewidth}
        \includegraphics[width=1\linewidth]{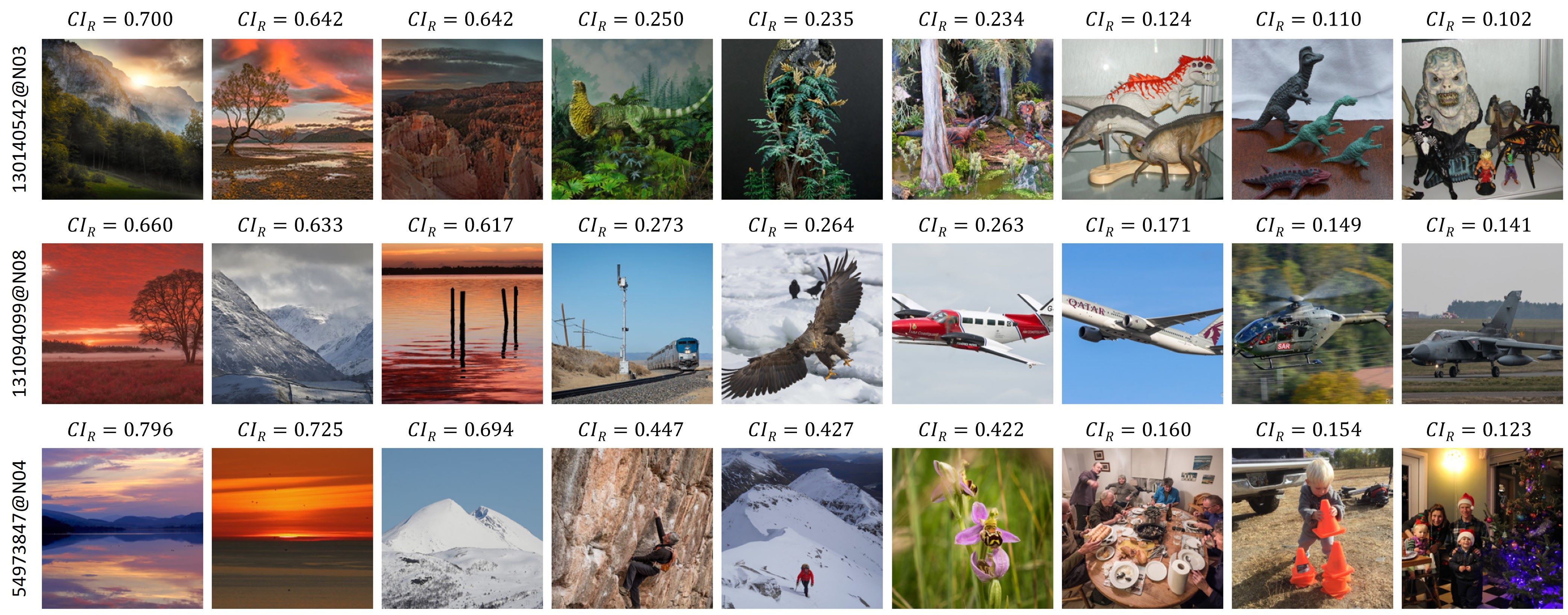}
        \caption{As expected, common interesting images show similarities across users, whereas their personal preferences differ. As the $CI_R$ score decreases, the subjectivity of interest also increases. For example, one user prefers dinosaur figures, while another prefers airplanes or family pictures.}
        \label{fig:userPred}
    \end{subfigure}

    \begin{subfigure}{\linewidth}
        \includegraphics[width=1\linewidth]{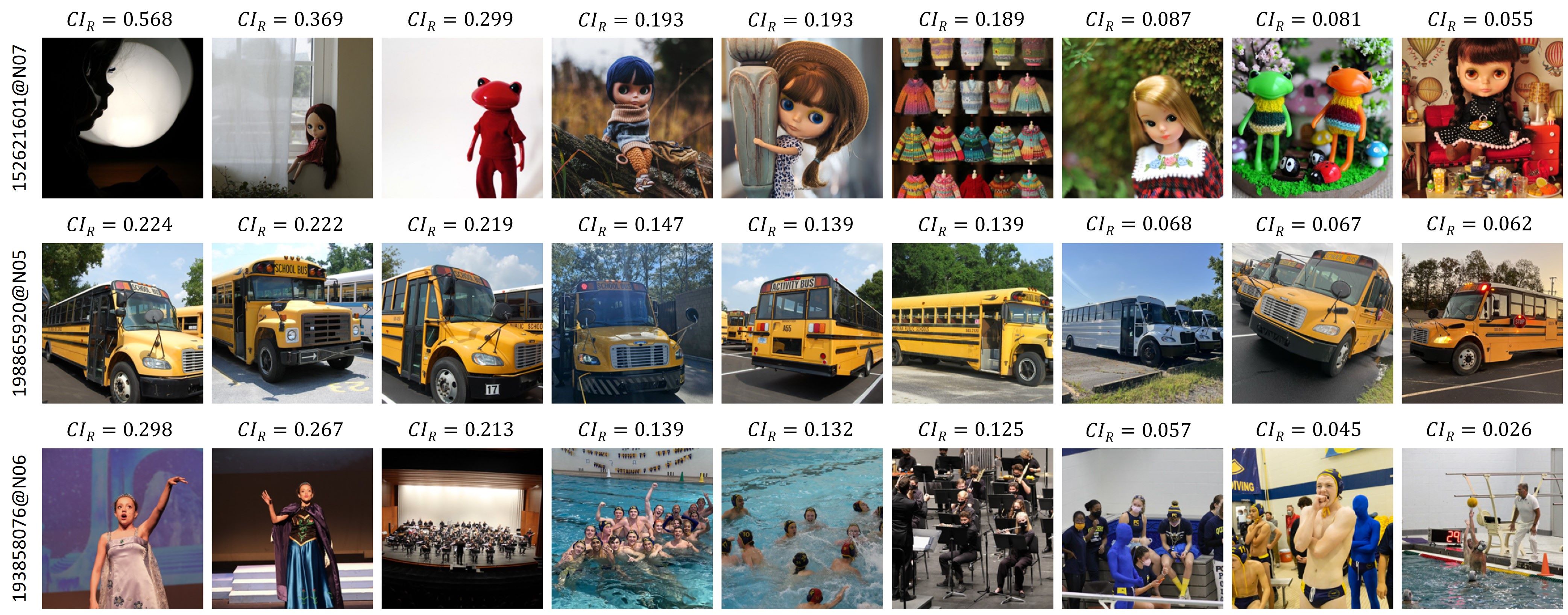}
        \caption{User with focus on very subjective images, including topics (\eg, personal preference) or personal events that are important to the particular user but not necessarily to the public.}
        \label{fig:userPredSub}
    \end{subfigure}

    \begin{subfigure}{\linewidth}
        \includegraphics[width=1\linewidth]{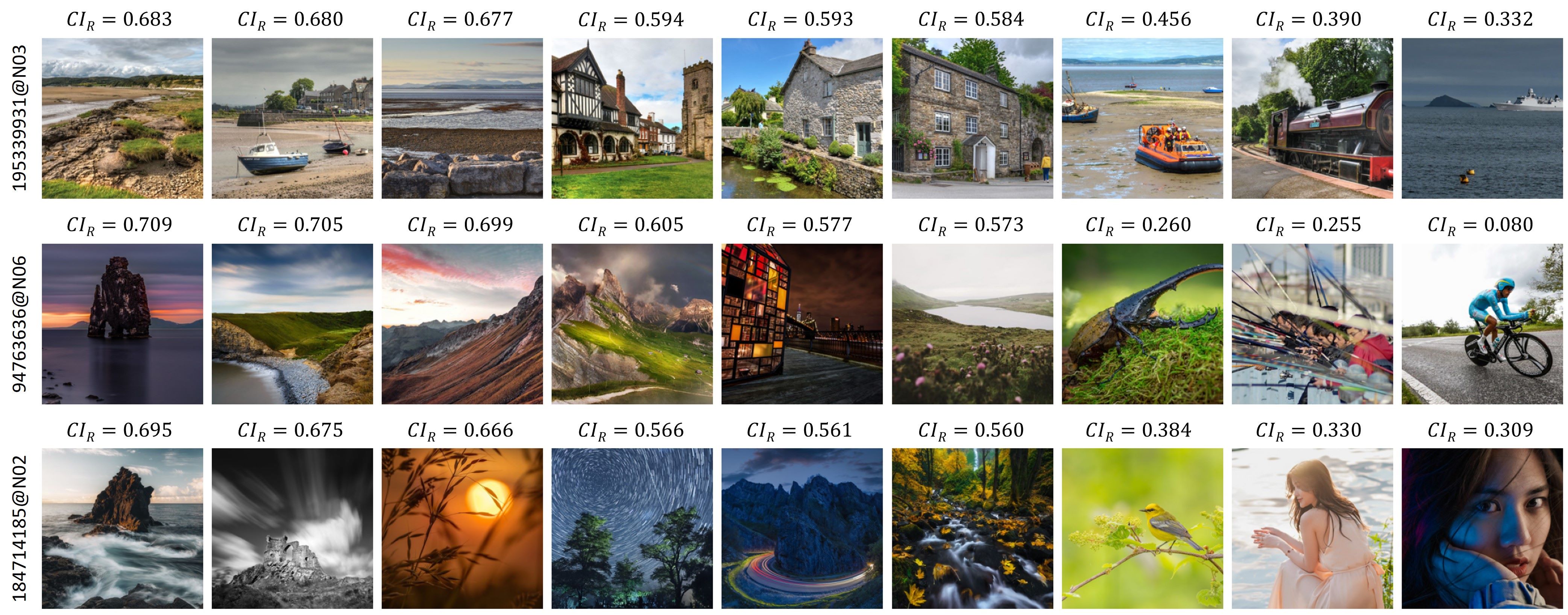}
        \caption{``Mainstream'' - users, mostly favoring commonly interesting images.}
        \label{fig:userPredCom}
    \end{subfigure}
    \caption{Most users have a mix of commonly interesting and more subjective images (a). Other users like very specific topics (b); some are mainstream and mostly like commonly interesting images (c). {\color{blue}{Further user examples from the test set are included in the supplementary material for a closer investigation.}}}
    \label{fig:userPredAll}
\end{figure}

\textbf{\emph{FlickrUser} Data.} We tested the model with users' favorite images not used for the definition of $CI$. Fig.~\ref{fig:userPred} shows common and subjectively interesting images from three different users. Looking at the first three images of each user, the images show a semantic similarity of aesthetic landscapes. The lower the $CI_R$ score of the images, the more subjective the interest in the images. On the one hand, figures of dinosaurs are more interesting for the first user; on the other hand, airplanes or family pictures are more interesting for the second or third user. Our experiments show that most users like a mix of common and subjective interesting images. Nevertheless, quite some users focus on specific topics or preferences images that seem to recall personal memories; see Fig.~\ref{fig:userPredSub}. Others focus on high-quality, commonly interesting images; see Fig.~\ref{fig:userPredCom}. However, all three user groups highlight the subjective nature of the topic.

\textbf{Ranking images of the same scene/ same object.} To test how our model generalizes on images that should represent the same object but differ in viewpoint or time of day. We used different images of the Eiffel Tower (Fig.~\ref{fig:eiffelTower}) and the Statue of Liberty  (Fig.~\ref{fig:statueOfLiberty}). The Eiffel Tower at sunset is of common interest. The Eiffel Tower, made of LEGO bricks or a selfie, on the other hand, is of subjective interest ($CI_R$ scores ranging from 0.211 to 0.606). Similar results are obtained for the other location.

\begin{figure}[t]
    \centering
    \begin{subfigure}{0.325\linewidth}
        \centering
        \includegraphics[width=1\linewidth]{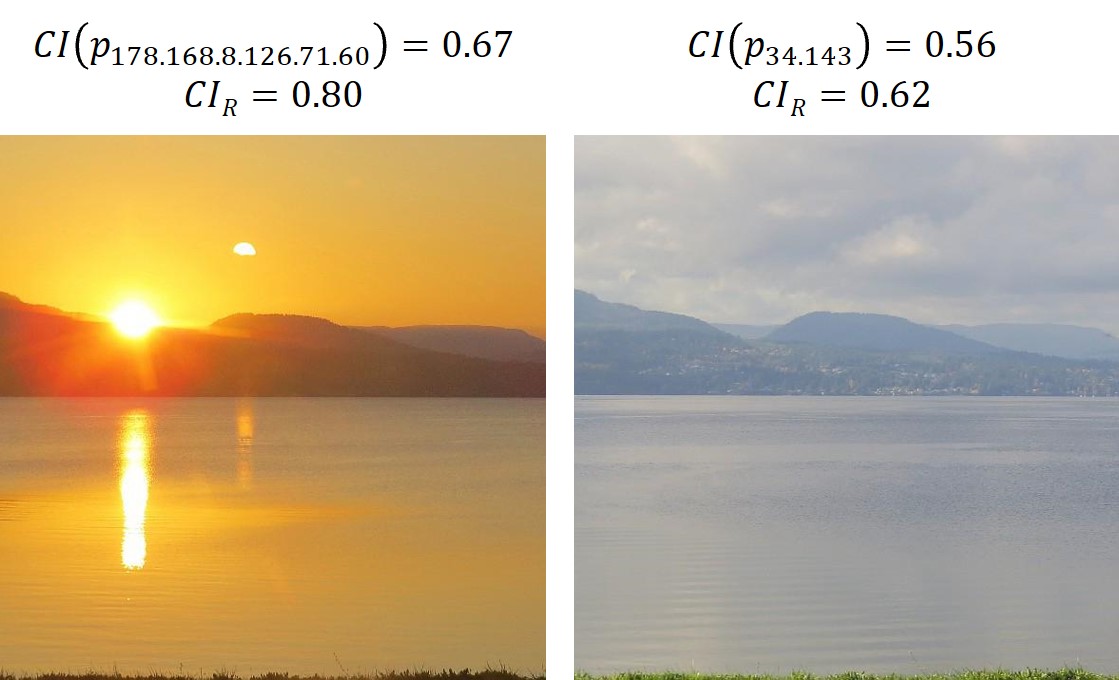}
        \caption{Same scene viewed by a webcam at different time of day}
        \label{fig:webcamsCompare}
    \end{subfigure}
    \hfill
    \begin{subfigure}{0.325\linewidth}
        \centering
        \includegraphics[width=1\linewidth]{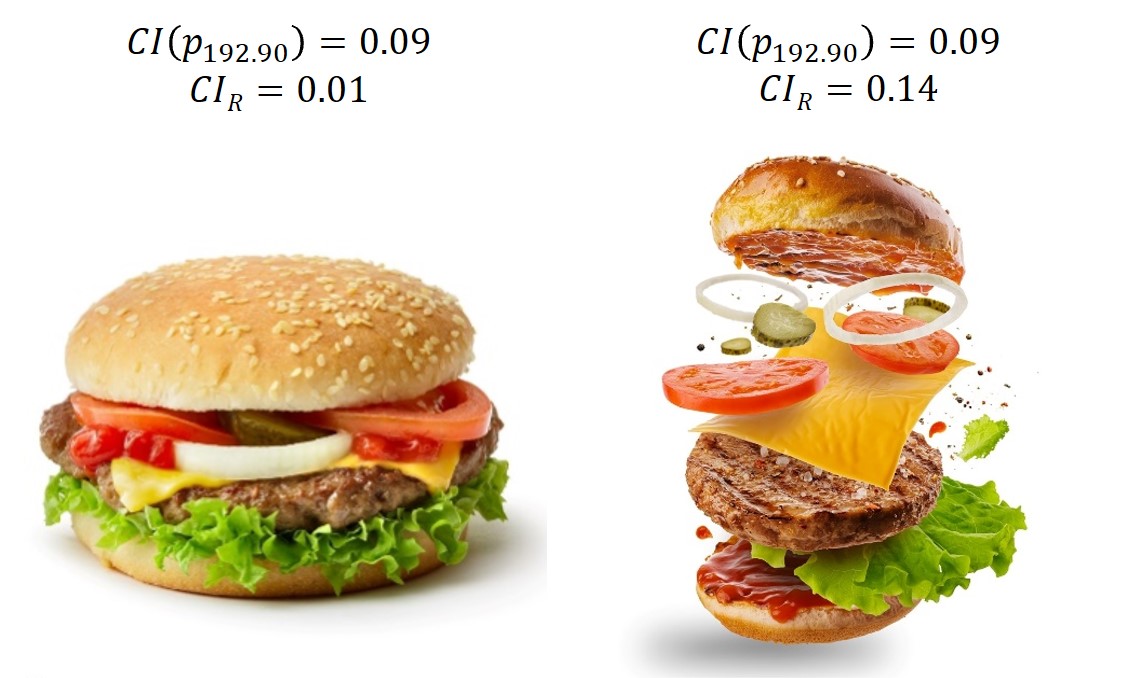}
        \caption{Marketing and Advertisement}
        \label{fig:compareBurger}
    \end{subfigure}
    \hfill
    \begin{subfigure}{0.325\linewidth}
        \centering
        \includegraphics[width=1\linewidth]{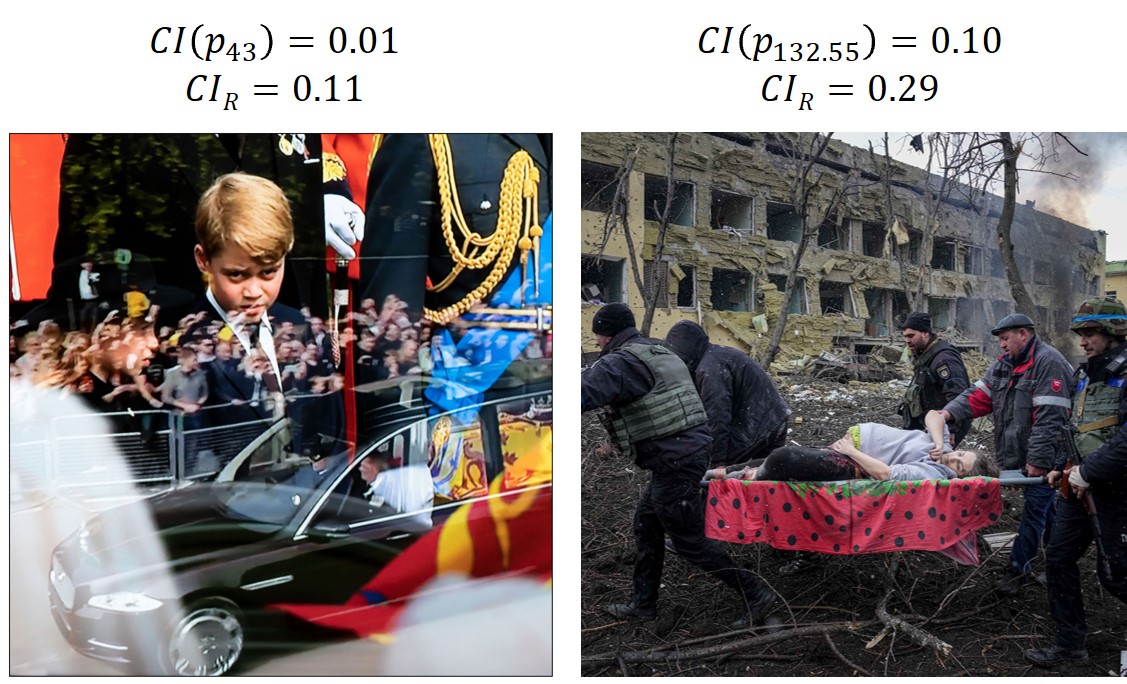}
        \caption{TIME's Top 100 Photos of 2022}
        \label{fig:compareTimes}
    \end{subfigure}
    \caption{Our model allows for selecting frames from webcams that are interesting to a wider audience (a) or ranking semantic similar images (b). However, it fails when other contextual information (top-down information) needs to be taken into account (c).}
\end{figure}

Let's consider, as another example, a static outdoor webcam \cite{patbaywebcam}. As shown in Fig.~\ref{fig:webcamsCompare}, an image with sunset has a much higher $CI$ and $CI_R$ score than the ``regular'' image, consistent with our analysis. Fig.~\ref{fig:compareBurger} illustrates two images of a burger. Both images have the same (low) $CI$ score. However, the fine-grained $CI_R$ score of the burger in motion is significantly higher than the static burger. This is in line with current research in the field of marketing~\cite{Grigsby2023}, which shows that images containing motion are more interesting than without.

\textbf{Limitations.} Fig.~\ref{fig:compareTimes} shows two images which made it into the \emph{TIMES Top 100 Images of 2022}\cite{timetop100}. So, both images might be considered interesting. However, due to the lack of top-down information (such as the context of being an image of the Russia-Ukraine war or the burial of Queen Elizabeth II), these images have quite low $CI_R$ scores as they are compared to similar semantic images, which usually rank low.

\section{Conclusion and Further Work}
\label{sec:conclusion}
Visual interestingness is a complex and multifaceted concept influenced by various factors. We focus on the subjective perception of observers. In contrast to other work, we presented a completely data-driven definition to score images as being either commonly or more subjectively interesting. We further determine image properties that make images more appealing to a broader audience. Some findings seem obvious in hindsight and are well aligned with related work (\eg, professionally taken and aesthetically pleasing images rank higher~\cite{gardezi2021whatmakes, Gygli2013TheImages, Constantin2021VisualReview}). Others might be more surprising, \eg, the presence of objects or persons alone is much more important for subjective interestingness. Finally, we trained a computational model, showing use cases and discussing limitations. Currently, only bottom-up information, solely obtained from the images, is used. Further work has to take the missing top-down information into account.

Additional insights might be gained by expanding beyond the Flickr domain -- a social photo-sharing platform. Firstly, our definition of \emph{common} interest still relies on a positive user commitment. Maybe not everything one finds interesting is worth a like (or one does not want to commit to it publicly). Secondly, some images might not even be uploaded due to legal, ethical, or other concerns. As shown in previous works~\cite{iaps2010, Gygli2013TheImages}, such images arouse interest but are based on negative stimuli.

{\small
{\bf Acknowledgements.} This research was funded by the Swiss National Science Foundation (SNSF) under grant number 206319 ``Visual Interestingness -- All images are equal but some images are more equal than others''.}


%
%
\bibliographystyle{splncs04}
\bibliography{main}
\end{document}